\documentclass[]{youtu} 

\usepackage{mathpazo}
\usepackage{graphicx}
\usepackage[numbers]{natbib}
\usepackage{enumitem}
\usepackage{graphicx}
\usepackage[most]{tcolorbox}
\usepackage{subcaption}

\usepackage{tcolorbox}
\usepackage{enumitem}
\usepackage{xcolor}
\usepackage{fontawesome}
\usepackage{multirow}

\definecolor{myborder}{RGB}{73, 86, 102}
\definecolor{myRed}{RGB}{240, 48, 159}
\definecolor{mylightblue}{RGB}{235, 245, 255}

\tcbuselibrary{skins}

\newcommand{\agentitem}[1]{\noindent\makebox[4em][l]{\textnormal{[Agent]}} \begin{minipage}[t]{\dimexpr\linewidth-4em\relax}#1\end{minipage}\\\smallskip}
\newcommand{\toolitem}[1]{\noindent\makebox[4em][l]{\textnormal{[Tool]}} \begin{minipage}[t]{\dimexpr\linewidth-4em\relax}#1\end{minipage}\\\smallskip}

\title{Training-Free Group Relative Policy Optimization}
\affiliation{Youtu-Agent Team$^*$}

\date{October 9, 2025}
\correspondence{tristanli@tencent.com}
\sourcecode{https://github.com/TencentCloudADP/youtu-agent/tree/training_free_GRPO}

\begin{document}

\abstract{Recent advances in Large Language Model (LLM) agents have demonstrated their promising general capabilities. However, their performance in specialized real-world domains often degrades due to challenges in effectively integrating external tools and specific prompting strategies. While methods like agentic reinforcement learning have been proposed to address this, they typically rely on costly parameter updates, for example, through a process that uses Supervised Fine-Tuning (SFT) followed by a Reinforcement Learning (RL) phase with Group Relative Policy Optimization (GRPO) to alter the output distribution. However, we argue that LLMs can achieve a similar effect on the output distribution by learning experiential knowledge as a token prior, which is a far more lightweight approach that not only addresses practical data scarcity but also avoids the common issue of overfitting. To this end, we propose Training-Free Group Relative Policy Optimization (Training-Free GRPO), a cost-effective solution that enhances LLM agent performance without any parameter updates. 
Our method leverages the group relative semantic advantage instead of numerical ones within each group of rollouts, iteratively distilling high-quality experiential knowledge during multi-epoch learning on a minimal ground-truth data. Such knowledge serves as the learned token prior, which is seamlessly integrated during LLM API calls to guide model behavior. Experiments on mathematical reasoning and web searching tasks demonstrate that Training-Free GRPO, when applied to DeepSeek-V3.1-Terminus, significantly improves out-of-domain performance.
With just a few dozen training samples, Training-Free GRPO outperforms fine-tuned small LLMs with marginal training data and cost.}
\maketitle

\begin{figure}[h]
    \centering
    \vspace{4mm}
    \includegraphics[width=1\linewidth]{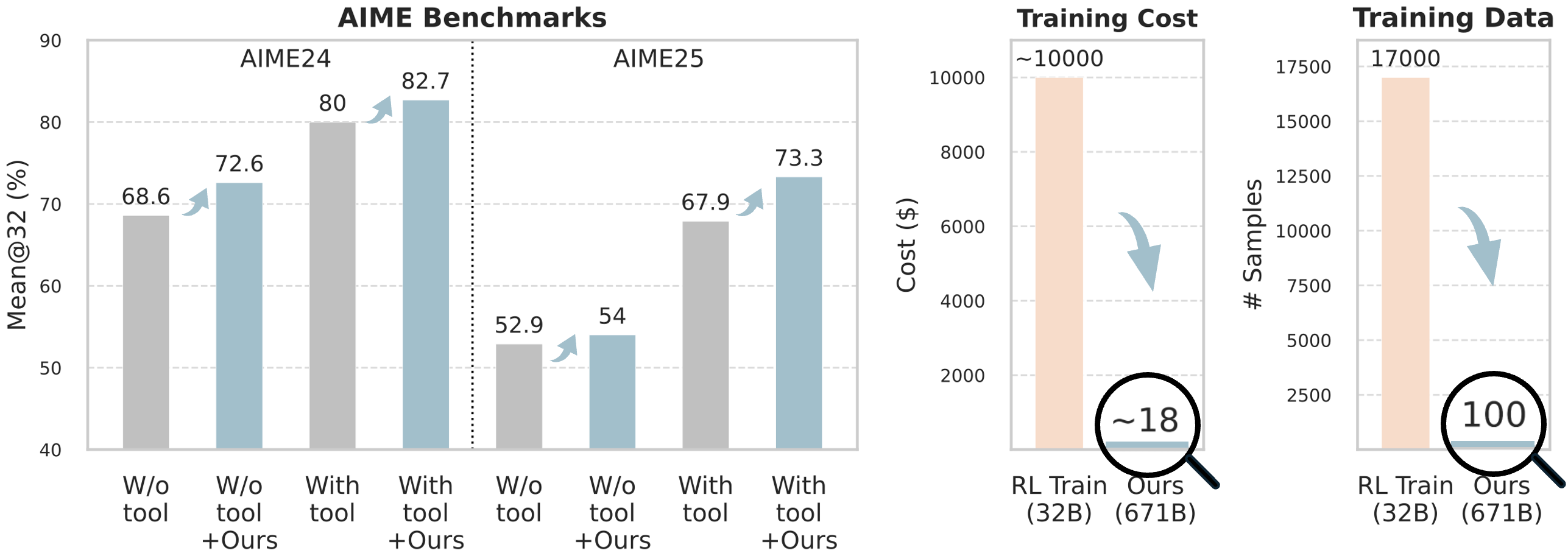}
    \caption{Applying Training-Free GRPO on both prompting (without tools) and ReAct~\citep{react} (with tools) achieve improved Mean@32 on AIME benchmarks~\citep{AIME} with DeepSeek-V3.1-Terminus~\citep{deepseekv3}. It consumes significantly fewer training data and lower costs on the 671B LLM than fine-tuning a 32B model~\citep{retool}, serving as a cost-effective alternative to RL methods.}
    \label{fig:spotlight}
\end{figure}

\renewcommand{\thefootnote}{*}
\footnotetext{Full author list in contributions.}
\renewcommand{\thefootnote}{\arabic{footnote}}

\vspace{-.1em}

\section{Introduction}
\label{sec:introduction}

Large Language Models (LLMs) are emerging as powerful general-purpose agents capable of interacting with complex, real-world environments. 
They have shown remarkable capabilities across a wide range of tasks, including complex problem-solving~\citep{retool, zerotir, simpletir}, advanced web research~\citep{tongyidr, webshaper, searchr1, mirothinker}, code generation and debugging~\citep{codeagent, agentcoder}, and proficient computer use~\citep{guiagents, mobileagent, textgrad}. 
Despite their impressive capabilities, LLM agents often underperform in specialized, real-world domains.
These scenarios typically demand the integration of external tools (e.g., calculators, APIs, databases), along with domain-specific task definitions and prompting strategies. 
Deploying a general-purpose agent out-of-the-box in such settings often results in suboptimal performance due to limited familiarity with domain-specific requirements or insufficient exposure to necessary tools.

To bridge this gap, agentic training has emerged as a promising strategy to facilitate the adaptation of LLM agents to specific domains and their associated tools~\citep{retool, tongyidr, webshaper, AFM}. 
Recent advancements in agentic reinforcement learning (Agentic RL) approaches have employed Group Relative Policy Optimization (GRPO)~\citep{grpo} and its variants~\citep{drgrpo, dapo, gspo} to align model behaviors in the parameter space.
Although these methods effectively enhance task-specific capabilities, their reliance on tuning LLM parameters poses several practical challenges:

\vspace{-5mm}
\begin{itemize}[leftmargin=*]
    \item \textbf{Computational Cost:} Even for smaller models, fine-tuning demands substantial computational resources, making it both costly and environmentally unsustainable. For larger models, the costs become prohibitive. Furthermore, fine-tuned models require dedicated deployment and are often limited to specific applications, rendering them inefficient for low-frequency use cases compared to more versatile general-purpose models.
    \item \textbf{Poor Generalization:} Models optimized via parameter tuning often suffer from unsatisfactory cross-domain generalization, limiting their applicability to narrow tasks. Consequently, multiple specialized models must be deployed to handle a comprehensive set of tasks, significantly increasing system complexity and maintenance overhead.
    \item \textbf{Data Scarcity:} Fine-tuning LLMs typically necessitates large volumes of high-quality, carefully annotated data, which are often scarce and prohibitively expensive to obtain in specialized domains.
    Additionally, with limited samples, models are highly susceptible to overfitting, leading to poor generalization.
    \item \textbf{Diminishing Returns:} The prohibitive training costs usually compel existing approaches to fine-tune smaller LLMs with fewer than 32 billion parameters, due to resource constraints rather than optimal design choices.
    While larger models would be preferred, the computational expense of fine-tuning necessitates this compromise. Paradoxically, API-based or open-source larger LLMs often deliver better cost-performance ratios through scalability and continuous model updates. However, these general-purpose models underperform in specialized domains where fine-tuning is necessary, creating a cost-performance dilemma.
\end{itemize}

Such limitations inherent in parameter tuning motivate a fundamental research question: \textit{Is applying RL in parametric space the only viable approach? Can we enhance LLM agent performance in a non-parametric way with lower data and computational costs?}

We answer this question affirmatively by proposing \textbf{Training-Free Group Relative Policy Optimization (Training-Free GRPO)}, a novel and efficient method that improves LLM agent behavior in a manner similar to vanilla GRPO, while preserving the original model parameters unchanged.
Our approach is motivated by the insight that LLMs already possess the fundamental capability to adapt to new scenarios, requiring only minimal practice through limited samples to achieve strong performance. Thus, instead of adapting their output distribution through parameter tuning, in-context learning~\citep{icl} that leverages a lightweight \textit{token prior} can also encapsulate experiential knowledge learned from a minimal training dataset.

Training-Free GRPO retains the multi-epoch learning mechanism of vanilla GRPO.
In each epoch, multiple outputs are generated to deliver a group of rollouts for every query, which helps to explore the policy space and evaluate potential strategies. While vanilla GRPO relies on gradient-based parameter updates to iteratively improve policy performance, Training-Free GRPO eliminates this requirement by employing inference-only operations using LLMs. 
At each optimization step, rather than calculating a numerical advantage for gradient ascent within each group of rollouts, our method leverages LLMs to introspect on each group and distill a semantic advantage. 
Such advantage refines external experiential knowledge and guide policy outputs based on evolving contextual priors, thereby achieving policy optimization effects without modifying any model parameters.

By evaluating challenging mathematical reasoning and interactive web searching tasks, we demonstrate that our method significantly enhances the performance of frozen policy models such as DeepSeek-V3.1-Terminus~\citep{deepseekv3} with only dozens of training samples. 
It surpasses fine-tuned 32B models in performance while requiring only a fraction of the computational resources, offering a simple and much more efficient alternative to traditional fine-tuning techniques.

Our principal contributions are threefold:
\begin{itemize}[leftmargin=*]
    \item \textbf{A New Training-Free RL Paradigm:} We introduce Training-Free GRPO, which shifts policy optimization from the parameter space to the context space by leveraging evolving experiential knowledge as token priors without gradient updates.
    \item \textbf{Semantic Group Advantage:} We replace numerical group advantage in vanilla GRPO with \textit{semantic group advantage}, enabling LLMs to introspect their own rollouts and continuously updating experiential knowledge at multiple optimization steps.
    \item \textbf{Data and Computational Efficiency:} Experiments confirm that Training-Free GRPO effectively enhances the performance of a frozen policy with minimal training samples, offering a practical and cost-effective alternative across different domains.
    \item \textbf{Superior Generalization:} By leaving model parameters frozen and plugging in different token priors, our approach fully preserves the generalization power, eliminating the cost and complexity of deploying multiple fine-tuned specialists.
\end{itemize}

\section{Training-Free GRPO}
\label{sec:method}

\begin{figure}[t]
    \centering
    \includegraphics[width=1\linewidth]{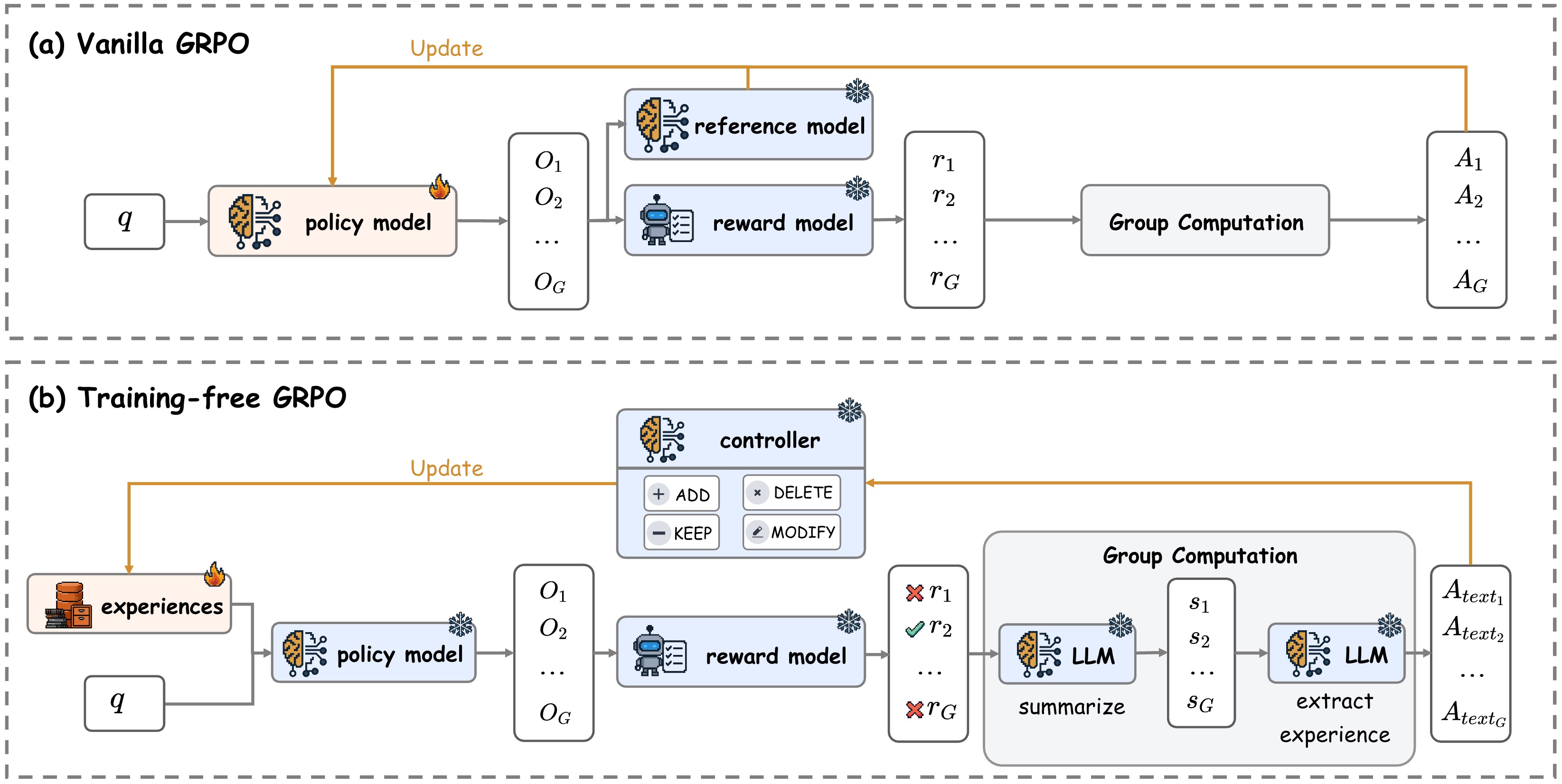}
    \caption{Comparison of vanilla GRPO and Training-Free GRPO.}
    \label{fig:compare}
\end{figure}

\begin{figure}[t]
    \centering
    \includegraphics[width=1\linewidth]{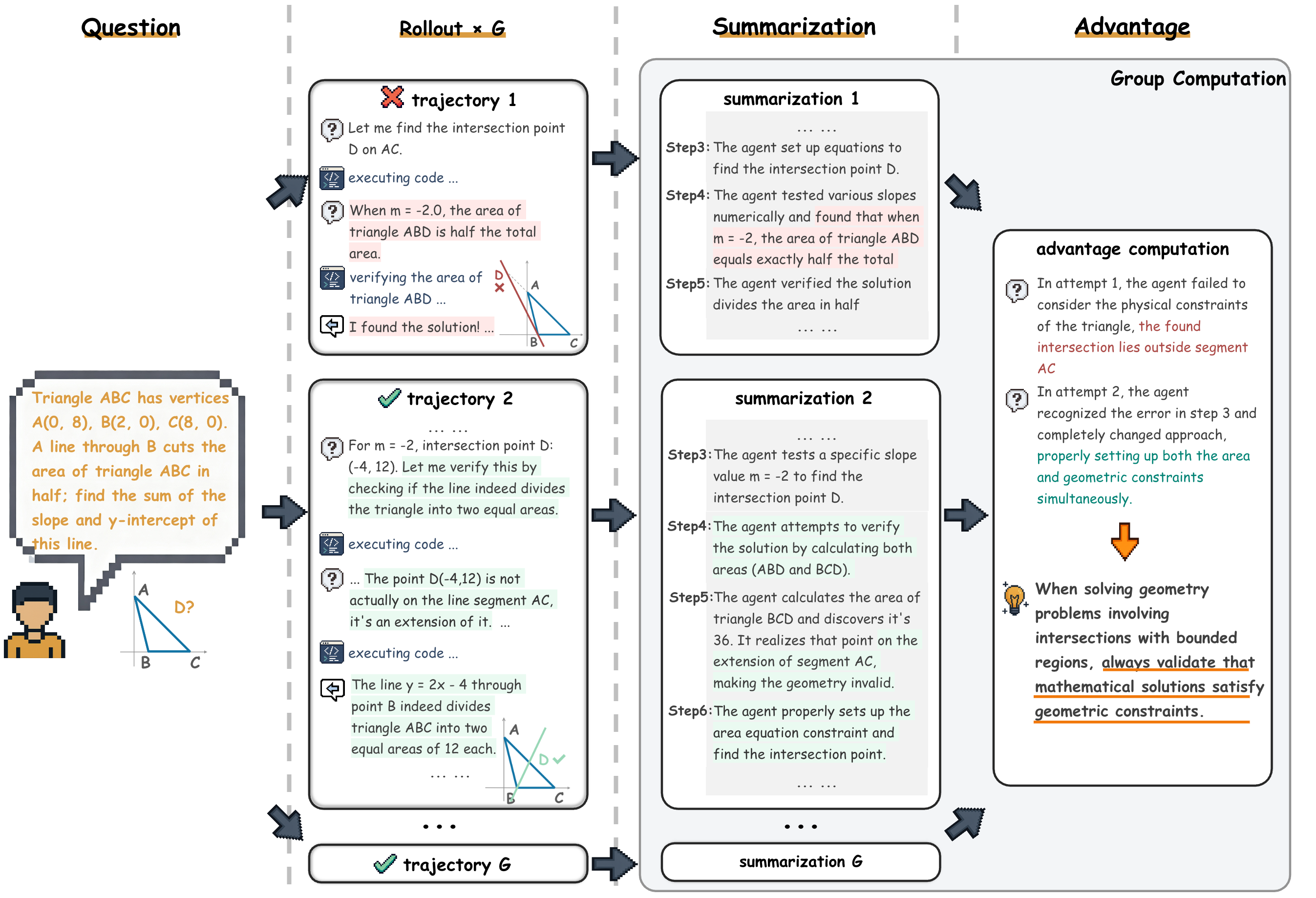}
    \caption{Example of a Training-Free GRPO learning step.}
    \label{fig:case}
\end{figure}

In this section, we introduce our Training-Free GRPO, a method designed to replicate the alignment benefits of the GRPO algorithm without performing any gradient-based updates to the policy model's parameters. 

\textbf{Vanilla GRPO.}
As shown in Figure~\ref{fig:compare}, the vanilla GRPO procedure operates by first generating a group of $G$ outputs $\{o_1, o_2, \dots, o_G\}$ for a given query $q$ using the current policy LLM $\pi_\theta$, i.e., $\pi_\theta(o_i \mid q)$. Each output $o_i$ is then independently scored with a reward model $\mathcal{R}$.
Subsequently, with rewards $\mathbf{r}=\{r_1, \dots, r_G\}$, it calculates a group-relative advantage $\hat{A}_i = \frac{r_i - \text{mean}(\mathbf{r})}{\text{std}(\mathbf{r})}$ for each output $o_i$. 
By combining a KL-divergence penalty against a reference model, it constructs a PPO-clipped objective function $\mathcal{J}_{\text{GRPO}}(\theta)$, which is then maximized to update the LLM parameters $\theta$. 

Training-Free GRPO repurposes the core logic of this group-based relative evaluation but translates it into a non-parametric, inference-time process. 
Instead of updating the parameters $\theta$, we leave $\theta$ permanently frozen and maintain an external \textit{experiential knowledge} $\mathcal{E}$, which is initialized to $\emptyset$.

\textbf{Rollout and Reward.}
As shown in Figure~\ref{fig:compare}, our rollout and reward process mirrors that of GRPO exactly.
Given a query $q$, we perform a parallel rollout to generate a group of $G$ outputs $\{o_1, o_2, \dots, o_G\}$ using the LLM. 
Notably, while GRPO uses the current trainable policy $\pi_\theta$, our policy conditions on the experiential knowledge, $\pi_\theta(o_i|q, \mathcal{E})$. 
Identical to the standard GRPO setup, we score each output $o_i$ by the reward model $\mathcal{R}$ to obtain a scalar reward $r_i = \mathcal{R}(q, o_i)$.

\textbf{Group Advantage Computation.}
To provide an optimization direction for policy parameters, vanilla GRPO computes a numerical advantage $\hat{A}_i$ that quantifies each output $o_i$'s relative quality within its group.
Similarly, Training-Free GRPO performs an analogous comparison within each group but produces a \textit{group relative semantic advantage} in the form of natural language experience, as shown in Figure~\ref{fig:case}.
Since $\hat{A}_i=0$ when all $G$ outputs receive identical rewards (i.e., $\text{std}(\mathbf{r})=0$) in vanilla GRPO, we generate such semantic advantages only for groups with both clear winners and losers. 
Specifically, for each output $o_i$, we first ask the same LLM $\mathcal{M}$ to provide a corresponding summary $s_i=\mathcal{M}(p_{\text{summary}}, q, o_i)$ separately, where $p_{\text{summary}}$ is a prompt template that incorporates the query $q$ and output $o_i$ to form a structured summarization request. 
Given the summaries $\{s_1, s_2, \dots, s_G\}$ and the current experiential knowledge $\mathcal{E}$, the LLM $\mathcal{M}$ articulates the reasons for the relative success or failure of the outputs, followed by extracting a concise natural language experience $A_{\text{text}}=\mathcal{M}({p_{\text{extract}}}, q, s_i, \mathcal{E})$, where $p_{\text{extract}}$ is another prompt template for experience extraction. 
This natural language experience $A_{\text{text}}$ serves as our semantic advantage, functionally equivalent to vanilla GRPO's $\hat{A}_i$, encoding the critical experiential knowledge of what actions lead to high rewards.

\textbf{Optimization.}
Whereas vanilla GRPO updates its model parameters $\theta$ via gradient ascent on $\mathcal{J}_{\text{GRPO}}(\theta)$ computed by all advantages in a single batch, we update our experience library $\mathcal{E}$ using all semantic advantages $A_{\text{text}}$ from the current batch. 
Specifically, given the existing experiences library $\mathcal{E}$, we prompt the LLM to generate a list of operations based on all these $A_{\text{text}}$, where each operation could be:
\vspace{-6mm}

\begin{itemize}[leftmargin=*]
\item \textit{Add:} Directly append the experience described in $A_{\text{text}}$ to the experience library $\mathcal{E}$.
\item \textit{Delete:} Based on $A_{\text{text}}$, remove a low-quality experience from the experience library $\mathcal{E}$.
\item \textit{Modify:} Refine or improve an existing experience in the experience library $\mathcal{E}$ based on insights from $A_{\text{text}}$.
\item \textit{Keep:} The experience library $\mathcal{E}$ remains unchanged.
\end{itemize}

\vspace{-6mm}
After updating the experience library $\mathcal{E}$, the conditioned policy $\pi_\theta(y|q, \mathcal{E})$ produces a shifted output distribution in subsequent batches or epochs. This mirrors the effect of a GRPO policy update by steering the model towards higher-reward outputs, but achieves this by altering the context rather than the model's fundamental parameters. The frozen base model $\pi_\theta$ acts as a strong prior, ensuring output coherence and providing a built-in stability analogous to the KL-divergence constraint in GRPO that prevents the policy from deviating excessively from $\pi_{\text{ref}}$.

\section{Evalution}
\label{sec:evaluation}

To compare Training-Free GRPO with competitive baselines, we conduct comprehensive experiments on both mathematical reasoning and web searching benchmarks.

\subsection{Mathematical Reasoning}
\label{sec: math results}

\textbf{Benchmarks.}
We conduct our evaluation on the challenging AIME24 and AIME25 benchmarks~\citep{AIME}, which are representative of complex, out-of-domain mathematical reasoning challenges. To ensure robust and statistically reliable results, we evaluate each problem with 32 independent runs and report the average Pass@1 score, which we denote as Mean@32. 

\textbf{Setup.}
We primarily focus on large powerful LLMs that are usually hard and expensive to be fine-tuned in real-world applications, such as DeepSeek-V3.1-Terminus~\citep{deepseekv3}.
We include two basic configurations: (1) Direct Prompting without tool use (a text-only input/output process), and (2) ReAct~\citep{react} with a code interpreter (CI) tool.
To apply Training-Free GRPO experiments, we randomly sample 100 problems from the DAPO-Math-17K dataset~\citep{dapo}, denoted as DAPO-100. We run the learning process for 3 epochs with a single batch per epoch (i.e., 3 steps), using a temperature of $0.7$ and a group size of $5$ during the learning phase.
For out-of-domain evaluation on AIME 2024 and 2025 benchmarks, we use a temperature of $0.3$.

\begin{table}[t]
\centering
\small
\caption{Mean@32 on AIME 2024 and AIME 2025 benchmarks (\%).}
\label{tab:aime large}
\begin{tabular}{lcclcll}
\toprule
\textbf{Method} & \textbf{Learning Cost} & \textbf{Training Set} & \textbf{Model} & \textbf{Tool} & \textbf{AIME24} & \textbf{AIME25} \\
\midrule
Direct & - & - & \textit{DeepSeek-V3.1-Terminus} & - & 68.6 & 52.9 \\
+ Training-Free GRPO & $\approx\$8$ & DAPO-100 & \textit{DeepSeek-V3.1-Terminus} & - & \textbf{72.6} \textcolor{youtuBlue}{\footnotesize ($\uparrow$4.0)} & \textbf{54.0} \textcolor{youtuBlue}{\footnotesize ($\uparrow$1.1)}\\
\midrule
ReAct~\cite{react} & - & - & \textit{DeepSeek-V3.1-Terminus} & CI & 80.0 & 67.9 \\
+ Training-Free GRPO & $\approx\$18$ & DAPO-100 & \textit{DeepSeek-V3.1-Terminus} & CI & \textbf{82.7} \textcolor{youtuBlue}{\footnotesize ($\uparrow$2.7)} & \textbf{73.3} \textcolor{youtuBlue}{\footnotesize ($\uparrow$ 5.4)}\\
\midrule
ReAct~\cite{react} & - & - & \textit{DeepSeek-V3.2-Exp} & CI & 71.0 & 61.8 \\
+ Training-Free GRPO & $\approx\$8$ & DAPO-100 & \textit{DeepSeek-V3.2-Exp} & CI & \textbf{73.1} \textcolor{youtuBlue}{\footnotesize ($\uparrow$2.1)} & \textbf{63.2} \textcolor{youtuBlue}{\footnotesize ($\uparrow$1.4)}\\
\bottomrule
\end{tabular}
\end{table}

\begin{figure}[t]
    \centering
    \includegraphics[width=\textwidth]{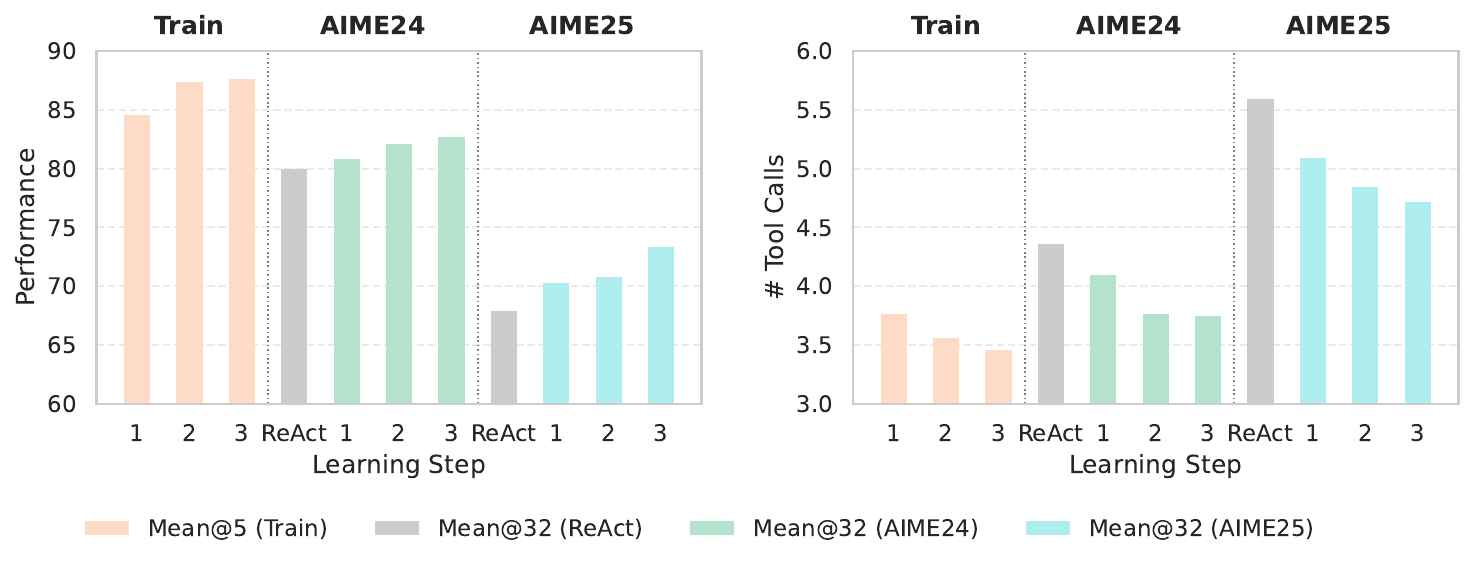}
    \caption{Statistics at each Training-Free GRPO step with tool use and DeepSeek-V3.1-Terminus.}
    \label{fig:curves}
\end{figure}

\textbf{Main Results.}
As illustrated in Table~\ref{tab:aime 32B}, Training-Free GRPO achieves substantial gains in mathematical reasoning, showing a clear advantage in performance across both the tool-use and non-tool-use scenarios. The strong baseline established by DeepSeek-V3.1-Terminus with ReAct~\citep{react} yields scores of $80.0\%$ on AIME24 and $67.9\%$ on AIME25. Critically, applying Training-Free GRPO to the frozen DeepSeek-V3.1-Terminus elevates its performance significantly, reaching $82.7\%$ on AIME24 and $73.3\%$ on AIME25. This represents a substantial absolute gain of $+2.7\%$ and $+5.4\%$, respectively, which is achieved with only $100$ out-of-domain training examples and zero gradient updates.
This performance surpasses various state-of-the-art RL methods like ReTool~\citep{retool} and AFM~\citep{AFM} trained on 32B LLMs (see Table~\ref{tab:aime 32B}), which typically require thousands of training examples and incur costs exceeding $\$10,000$. In contrast, Training-Free GRPO utilizes only $100$ data points with an approximate cost of $\$18$.
Such outcome suggests that in real-world applications, guiding a powerful but frozen model through context-space optimization is more effective and efficient than exhaustively fine-tuning a less capable model in parameter space.
For quantitative evaluation and learned experiences of Training-Free GRPO, please refer to Appendices~\ref{app: case study} and \ref{app: learned experiences}, respectively.

\textbf{Learning Dynamics.} As shown in Figure~\ref{fig:curves}, during the 3-step learning process, we observe a steady and significant improvement in Mean@5 on the training set. 
Concurrently, the Mean@32 performance on both AIME24 and AIME25 also improves with each step, demonstrating that the learned experiences from only $100$ problems generalize effectively and the necessity of multi-step learning.
Also, the average number of tool calls decreases during both training and out-of-domain evaluation on AIME benchmarks.
This suggests that Training-Free GRPO not only encourages correct reasoning and action, but also teaches the agent to use tools more efficiently and judiciously.
The learned experiential knowledge helps the agent to discover some shortcuts and avoid erroneous or redundant tool calls, validating the effectiveness of our semantic advantage guided optimization.

\begin{table}[t]
\centering
\small
\caption{Ablation study on DeepSeek-V3.1-Terminus (Mean@32, \%).}
\vspace{-1mm}
\label{tab: aime2425 ablation}
\setlength{\tabcolsep}{3.4mm}{
\begin{tabular}{lccc}
\toprule
\textbf{Method} & \textbf{Training Set} & \textbf{AIME24} & \textbf{AIME25} \\
\midrule
ReAct~\citep{react} & - & 80.0 & 67.9  \\
ReAct~\citep{react} + Directly Generated Experiences & - & 79.8 & 67.3 \\
ReAct~\citep{react} + Training-Free GRPO {\footnotesize(w/o ground truths)} & DAPO-100 & 80.7 & 68.9 \\
ReAct~\citep{react} + Training-Free GRPO {\footnotesize(w/o group computation)} & DAPO-100  & 80.4 & 69.3 \\
ReAct~\citep{react} + Training-Free GRPO & DAPO-100 & \textbf{82.7} & \textbf{73.3} \\
\bottomrule
\end{tabular}
}
\end{table}

\begin{table}[t]
\centering
\small
\vspace{2mm}
\caption{Mean@32 with smaller LLMs on AIME 2024 and AIME 2025 benchmarks (\%).}
\vspace{-1mm}
\label{tab:aime 32B}
\begin{tabular}{lclcll}
\toprule
\textbf{Method} & \textbf{Learning Cost} & \textbf{Model} & \textbf{Tool} & \textbf{AIME24} & \textbf{AIME25} \\
\midrule
ReAct~\citep{react} & - & \textit{Qwen2.5-32B-Instruct} & CI & 29.6 & 23.1 \\
ZeroTIR~\citep{zerotir} & $\approx\$20,000$ & \textit{Qwen2.5-32B-Instruct}& CI & 56.7 &  33.3 \\
SimpleTIR~\citep{simpletir} & $\approx\$20,000$ & \textit{Qwen2.5-32B-Instruct}& CI & 59.9 & 49.2 \\
ReTool~\citep{retool} & $\approx\$10,000$ & \textit{Qwen2.5-32B-Instruct}& CI & 67.0 & 49.3 \\
AFM~\citep{AFM} & $\approx\$10,000$ & \textit{Qwen2.5-32B-Instruct}& CI & 66.7 & 59.8 \\
\midrule
ReAct~\cite{react} & - & \textit{Qwen3-32B (Non-Thinking)} & CI & 29.1 & 19.5 \\
+ Training-Free GRPO & $\approx \$4$ & \textit{Qwen3-32B (Non-Thinking)} & CI & 33.5 \textcolor{youtuBlue}{\footnotesize ($\uparrow$4.4)} & 25.4 \textcolor{youtuBlue}{\footnotesize ($\uparrow$5.9)} \\
\midrule
ReAct~\cite{react} & - & \textit{Qwen2.5-72B-Instruct} & CI & 13.5 & 9.6 \\
+ Training-Free GRPO & $\approx \$3$ & \textit{Qwen2.5-72B-Instruct} & CI & 14.9 \textcolor{youtuBlue}{\footnotesize ($\uparrow$1.4)} & 11.4 \textcolor{youtuBlue}{\footnotesize ($\uparrow$1.8)} \\
\midrule
ReAct~\cite{react} & - & \textit{DeepSeek-V3.1-Terminus} & CI & 80.0 & 67.9 \\
+ Training-Free GRPO & $\approx\$18$ & \textit{DeepSeek-V3.1-Terminus} & CI & 82.7 \textcolor{youtuBlue}{\footnotesize ($\uparrow$2.7)} & 73.3 \textcolor{youtuBlue}{\footnotesize ($\uparrow$ 5.4)}\\
\bottomrule
\end{tabular}
\end{table}

\textbf{Effectiveness of Learned Experiences.} In Table~\ref{tab: aime2425 ablation}, we also include the ReAct enhanced with the experiences directly generated by DeepSeek-V3.1-Terminus, matching the quantity learned from Training-Free GRPO.
However, such directly generated experiences fail to improve the performance, highlighting the effectiveness of learned experiential knowledge from Training-Free GRPO.

\textbf{Robustness to Reward Signal.} Table~\ref{tab: aime2425 ablation} also presents a variant of Training-Free GRPO, where the ground truth answers are not provided during learning process. In such cases, the semantic advantage is directly obtained by comparing the rollouts within each group, where the LLM can only rely on implicit majority voting, self-discrimination and self-reflection to optimize the experiences. Although it does not surpass the default version with ground truths, Training-Free GRPO still achieves an impressive results of 80.7\% on AIME24 and 68.9\% on AIME25. It shows the robustness and applicability to domains where ground truths are scarce or unavailable, further broadening its practical utility.

\textbf{Removing Group Computation.} We also remove the group computation by setting the group size of $1$ in Training-Free GRPO, where the LLM can only distill semantic advantage from the single rollout of each query. The results in Table~\ref{tab: aime2425 ablation} show that it significantly harm the performance comparing with the default group size of $5$. It confirms the necessity of group relative computation, which enables the LLM to compare different trajectories within each group for better semantic advantage and experience optimization.

\textbf{Applicability to Different Model Sizes.}
By applying Training-Free GRPO on smaller LLMs, specifically Qwen3-32B~\citep{qwen3} and Qwen2.5-72B-Instruct~\citep{qwen25} with DAPO-100 dataset, we observe consistent improvements on the out-of-domain AIME benchmarks in Table \ref{tab:aime 32B}. Training-Free GRPO requires significantly fewer data and much lower learning cost, contrasting sharply with recent RL methods like ZeroTIR~\citep{zerotir}, SimpleTIR~\citep{simpletir}, ReTool~\citep{retool}, and AFM~\citep{AFM}, which often necessitate thousands of data points and substantial computational resources for parameter tuning. Furthermore, powered by larger models like DeepSeek-V3.1-Terminus, our approach achieves much higher Mean@32 on AIME benchmarks than all RL-trained models, while only incurring about $\$18$ for the learning process.

\subsection{Web Searching}

\begin{table}[t]
\small
\caption{Pass@1 on WebWalkerQA (\%).}
\vspace{-1mm}
\label{tab:webwalker_full}
\centering
\setlength{\tabcolsep}{8.6mm}{
\begin{tabular}{lcll}
\toprule
\textbf{Method} & \textbf{Training Set} & \textbf{Model} & \textbf{pass@1} \\
\midrule
ReAct~\citep{react}    &   -          & \textit{DeepSeek-V3.1-Terminus}    & 63.2\\
+ Training-Free GRPO & AFM-100 & \textit{DeepSeek-V3.1-Terminus}    & \textbf{67.8} \textcolor{youtuBlue}{\footnotesize ($\uparrow$4.6)} \\
\bottomrule
\end{tabular}
}
\end{table}

In this section, we evaluate the effectiveness of Training-Free GRPO in addressing web searching tasks by leveraging minimal experiential data to enhance agent behavior.

\textbf{Datasets.}
For training, we constructed a minimal training set by randomly sampling 100 queries from the AFM (Chain-of-Agents) web interaction RL dataset~\citep{AFM}, denoted as AFM-100. AFM provides high-quality, multi-turn interactions between agents and web environments, collected via reinforcement learning in realistic browsing scenarios. For evaluation, we employ WebWalkerQA benchmark~\citep{WebWalekrQA}, a widely-used dataset for assessing web agent performance. Its tasks require understanding both natural language instructions and complex web page structures, making it a rigorous evaluation framework for generalist agents.

\textbf{Methods.}
Our proposed Training-Free GRPO is applied to DeepSeek-V3.1-Terminus without any gradient-based updates. We perform 3 epochs of training-free optimization with a group size of $G=3$. The temperature settings follow those used in prior mathematical experiments.

\textbf{Main Results.}
We evaluate the effectiveness of our proposed Training-Free GRPO method on the WebWalkerQA benchmark. As shown in Table~\ref{tab:webwalker_full}, our method achieves a pass@1 score of 67.8\% using DeepSeek-V3.1-Terminus, a significant improvement over the baseline of 63.2\%. This result indicates that our approach effectively steers model behavior by leveraging learned experiential knowledge, surpassing the capabilities of static prompting ReAct strategy.

\begin{table}[t]
\small
\vspace{2mm}
\caption{Ablation results on WebWalkerQA subset (\%).}
\label{tab:webwalker_ablation}
\vspace{-1mm}
\centering
\begin{tabular}{lclcc}
\toprule
\textbf{Method} & \textbf{Training Set} & \textbf{Model} & \textbf{pass@1} & \textbf{pass@3} \\
\midrule
ReAct~\citep{react}    & -                              & \textit{QwQ-32B}         & 27.5 & 43.1 \\
ReAct~\citep{react} + Training-Free GRPO       & AFM-100              & \textit{QwQ-32B}         & 25.5 & 45.1 \\
\midrule
ReAct~\citep{react}         & -                         & \textit{DeepSeek-V3.1-Terminus}   & 66.7 & 74.5 \\
ReAct~\citep{react} + Directly Generated Experiences  & AFM-100   & \textit{DeepSeek-V3.1-Terminus}   & 64.7 & 76.5 \\
ReAct~\citep{react} + Training-Free GRPO {\footnotesize(w/o ground truths)} & AFM-100 & \textit{DeepSeek-V3.1-Terminus}   & 66.7 & 78.4 \\
ReAct~\citep{react} + Training-Free GRPO               & AFM-100      & \textit{DeepSeek-V3.1-Terminus}   & \textbf{68.6} & \textbf{78.4} \\
\bottomrule
\end{tabular}
\end{table}

\textbf{Ablation.}
We conduct ablation studies on a stratified random sample of 51 instances from the WebWalkerQA test set, where the sampling is proportionally stratified by difficulty level to ensure balanced representation across different levels of complexity. All ablated models are evaluated after 2 epochs of experience optimization. The results are summarized in Table \ref{tab:webwalker_ablation}.

Using directly generated experiences slightly degrades over ReAct (64.7\% vs. 66.7\% pass@1), confirming that mere in-context examples without proper optimization may not yield gains. Training-Free GRPO without ground truth maintains the same pass@1 as ReAct (66.7\%) but improves pass@3 to 78.4\%, demonstrating that relative reward evaluation can enhance consistency even without ground truth. The full Training-Free GRPO achieves the best performance (68.6\% pass@1 and 78.4\% pass@3), highlighting the importance of combining ground truth guidance with semantic advantage and experience optimization.

Applying Training-Free GRPO to QwQ-32B~\citep{qwq32b} yields only 25.5\% pass@1, significantly lower than the 66.7\% achieved with DeepSeek-V3.1-Terminus, and even under performing its own ReAct baseline (27.5\%). This may suggest that the effectiveness of our method is dependent on the underlying model’s reasoning and tool-use capabilities in complex tool use scenarios, indicating that model capability is a prerequisite for effective experience-based optimization.

\section{Comparing RL Learning on Context Space and Parameter Space}

\subsection{Cross-domain Transfer Analysis}
\label{sec:cross_domain}

\begin{table}[t]
\centering
\caption{Cross-domain transferability (Averaged pass@1, \%).}
\label{tab:cross_domain}
\begin{tabular}{lcccc}
\toprule
\multirow{2}{*}{\textbf{Method}} & \multirow{2}{*}{\textbf{Learned Domain}} & \multicolumn{2}{c}{\textbf{Math Reasoning}} & \textbf{Web Searching} \\
\cmidrule(lr){3-4} \cmidrule(lr){5-5}
 &  & AIME24  & AIME25 & WebWalker \\
\midrule
ReAct~\citep{react} \textit{(Qwen2.5-32B-Instruct)} & - & 29.6 & 23.1 & 31.9 \\
\midrule
ReTool~\citep{retool} \textit{(Qwen2.5-32B-Instruct)} & Math & 67.0 & 49.3 & 18.3 \\
MiroThinker~\citep{mirothinker} \textit{(Qwen3-32B)} & Web & 43.5 & 36.8 & 53.6 \\
Training-Free GRPO \textit{(DeepSeek-V3.1-Terminus)} & Math / Web & \textbf{82.7} & \textbf{73.3} & \textbf{67.8} \\
\bottomrule
\end{tabular}
\end{table}

A critical strength of Training-Free GRPO lies in its ability to achieve strong performance across diverse domains without suffering from the domain specialization trade-off observed in parameter-tuned methods. As demonstrated in Table~\ref{tab:cross_domain}, we observe the unsatisfactory performance when domain-specialized models are transferred to different domains. For instance, ReTool~\citep{retool} specifically trained on mathematical reasoning tasks, achieves competitive performance on AIME24 and AIME25 within its specialized domain. However, when transferred to web searching tasks on WebWalker, its performance drops dramatically to only $18.3\%$, which is much lower than ReAct~\citep{react} without fine-tuning. Similarly, though optimized for web interactions, MiroThinker \citep{mirothinker} significantly underperforms ReTool that is trained in the math domain on the AIME benchmarks. Such phenomenon highlights that parameter-based specialization narrows the model's capabilities to excel in the training domain at the expense of generalizability. In contrast, Training-Free GRPO applied to the frozen LLM achieves state-of-the-art performance in both domains simultaneously by simply plugging in domain-specific learned experiences.
Such cross-domain robustness makes Training-Free GRPO particularly valuable for real-world applications where agents must operate in multifaceted environments with diverse requirements.

\subsection{Computational Costs}

As shown in Figure~\ref{fig:spotlight}, we further analyze the economic benefits of Training-Free GRPO by comparing its computational costs with a vanilla RL training approach, specifically ReTool~\citep{retool}, on mathematical problem-solving tasks. This comparison highlights the practical benefits of our method in scenarios characterized by limited data, constrained budgets, or volatile inference demand.

\textbf{Training Cost.}
By replicating the training process of ReTool~\citep{retool} on Qwen2.5-32B-Instruct~\citep{qwen25}, we find that it requires approximately 20,000 $\times$ GPU hours with the rental price of \$0.5 per GPU hour, resulting in the total training expense amounts to roughly \$10,000. In contrast, Training-Free GRPO, when applied to DeepSeek-V3.1-Terminus, achieves superior performance on the AIME benchmarks (Table~\ref{tab:aime 32B}) while requiring only minimal costs.
It requires only $3$ training steps over $100$ samples completed within $6$ hours, which consumes 38M input tokens and 6.6M output tokens, amounting to a total cost of approximately \$18 based on the official DeepSeek AI pricing\footnote{\url{https://api-docs.deepseek.com/quick_start/pricing}. Most of the input tokens qualify for the lower cache hit pricing with ReAct framework, as each processing step typically involves reusing extensive prior context.}. The drastic reduction in training cost by over two orders of magnitude makes our approach especially cost-effective.

\textbf{Inference Cost.}
Deploying a trained model like ReTool-32B entails significant fixed infrastructure costs. In a typical serving setup with 4 $\times$ GPUs at the price of \$0.5 per GPU-hour,  vLLM-based batching requests can process about 400 problems per hour from the AIME benchmarks. The inference cost per problem averages \$0.005. While this per-instance cost is relatively low, it presupposes continuous GPU availability, which becomes inefficient under fluctuating or low request volumes. In contrast, Training-Free GRPO incurs a token-based cost.
On average, each request consumes 60K input tokens and 8K output tokens, totaling about \$0.02 per problem with cache hit pricing for reused contexts. 
Although per-query inference with a large API-based model is more expensive than with a dedicated small model, many real-world applications, particularly specialized or low-traffic services, experience irregular and modest usage patterns. In such cases, maintaining a dedicated GPU cluster is economically unjustifiable. By leveraging the shared, on-demand infrastructure of large model services like DeepSeek, Training-Free GRPO eliminates fixed serving overhead and aligns costs directly with actual usage. This pay-as-you-go model is distinctly advantageous in settings where demand is unpredictable or sparse.

\section{Related Work}
\label{sec:related_work}

\textbf{LLM Agents.}
By leveraging external tools, Large Language Models (LLMs) can overcome inherent limitations, such as lacking real-time knowledge and precise computation. This has spurred the development of LLM agents that interleave reasoning with actions. Foundational frameworks like ReAct~\citep{react} prompt LLMs to generate explicit chain-of-thought (CoT) and actionable steps, enabling dynamic planning through tool use. Furthermore, Toolformer~\citep{toolformer} demonstrates that LLMs can learn to self-supervise the invocation of APIs via parameter fine-tuning. Building on these principles, subsequent research has produced sophisticated single- and multi-agent systems, such as MetaGPT~\citep{metagpt}, CodeAct~\citep{codeact}, and OWL~\citep{owl}, which significantly enhance the quality of planning, action execution, and tool integration.

\textbf{Reinforcement Learning.}
Reinforcement learning (RL) has proven highly effective for aligning LLMs with complex and long-horizon goals. Foundational algorithms like Proximal Policy Optimization (PPO)~\citep{ppo} employ a policy model for generation and a separate critic model to estimate token-level value. Group Relative Policy Optimization (GRPO)~\citep{grpo} eliminates the need for a critic by estimating advantages directly from groups of responses. Recent research try to apply RL to transform LLMs from passive generators into autonomous agents that learn through environmental interaction. GiGPO~\citep{gigpo} implements a two-level grouping mechanism for trajectories, enabling precise credit assignment at both the episode and individual step levels. ReTool~\citep{retool} uses PPO to train an agent to interleave natural language with code execution for mathematical reasoning. Chain-of-Agents~\citep{AFM} facilitates multi-agent collaboration within a single model by using dynamic, context-aware activation of specialized tool and role-playing agents. Furthermore, Tongyi Deep Research~\citep{tongyidr} introduces synthetic data generation pipeline and conduct customized on-policy agentic RL framework.
However, such parameter-updating approaches result in prohibitive computational cost, which typically restricts application to LLMs with fewer than 32B parameters. Moreover, they only achieve diminishing returns compared to simply using larger, more powerful frozen LLMs. In contrast, our proposed Training-Free GRPO method seeks to achieve comparable or even better performance on state-of-the-art LLMs without any parameter updates, drastically reducing both data and computational requirements.

\textbf{Training-Free Methods.}
A parallel line of research aims to improve LLM behavior at inference time without updating model weights. The general approach is in-context learning (ICL)~\citep{icl}, which leverages external or self-generated demonstrations within a prompt to induce desired behaviors. More recent methods introduce iterative refinement mechanisms. Self-Refine~\citep{selfrefine} generates an initial output and then uses the same LLM to provide verbal feedback for subsequent revisions. Similarly, Reflexion~\citep{reflexion} incorporates an external feedback signal to prompt the model for reflection and a new attempt. In-context reinforcement learning (ICRL)~\citep{icrl1,icrl2} demonstrates that LLMs can learn from scalar reward signals by receiving prompts containing their past outputs and associated feedback. TextGrad~\citep{textgrad} proposes a more general framework, treating optimization as a process of back-propagating textual feedback through a structured computation graph. A key characteristic of these methods is their focus on iterative, within-sample improvement for a single query. In contrast, our Training-Free GRPO more closely mirrors traditional RL by learning from a separate dataset across multiple epochs to iteratively refine a shared, high-quality experience library for out-of-domain queries. Furthermore, given each query, unlike self-critique or context updates for a single trajectory, our method explicitly compares multiple rollouts per query for a semantic advantage to compare different trajectories in each group, which has been confirmed effective in Section~\ref{sec: math results}.
Specifically in the context for optimizing agent systems, Agent KB~\citep{agentkb} constructs a shared, hierarchical knowledge base to enable the reuse of problem-solving experiences across tasks. Unlike the complex reason-retrieve-refine process of Agent KB, Training-Free GRPO simply injects the learned experiences into the prompt. 
Moreover, Agent KB relies on hand-crafted examples and employs an off-policy learning paradigm only once by collecting trajectories in the different way of online inference. 
In contrast, our Training-Free GRPO uses a consistent pipeline and more closely mirrors on-policy RL with multi-epoch learning.

\section{Conclusion}
In this paper, we introduced Training-Free GRPO, a novel paradigm that shifts RL policy optimization from the parameter space to the context space. By leveraging group-based rollouts to iteratively distill a semantic advantage into an evolving experiential knowledge which serves as the token prior, our method successfully steers the output distribution of a frozen LLM agent, achieving significant performance gains in specialized domains. Experiments demonstrate that Training-Free GRPO not only surmounts the practical challenges of data scarcity and high computational cost but also outperforms traditional parameter-tuning methods. Our work establishes a new, highly efficient pathway for adapting powerful LLM agents, making advanced agentic capabilities more accessible and practical for real-world applications.

\newpage

\section*{Contributions}

\textbf{Authors} \quad
Yuzheng Cai\textsuperscript{\rm 1,2*} \quad Siqi Cai\textsuperscript{\rm 1*}\quad Yuchen Shi\textsuperscript{\rm 1*}\quad Zihan Xu\textsuperscript{\rm 1*} \quad Lichao Chen\textsuperscript{\rm 1,3} \quad Yulei Qin\textsuperscript{\rm 1} \quad Xiaoyu Tan\textsuperscript{\rm 1}\quad Gang Li\textsuperscript{\rm 1} \quad Zongyi Li\textsuperscript{\rm 1} \quad Haojia Lin\textsuperscript{\rm 1} \quad Yong Mao\textsuperscript{\rm 1} \quad Ke Li\textsuperscript{\rm 1\faEnvelopeO} \quad Xing Sun\textsuperscript{\rm 1}

\textbf{Affiliations} \quad
\textsuperscript{\rm 1}Tencent Youtu Lab\quad \textsuperscript{\rm 2}Fudan University\quad \textsuperscript{\rm 3}Xiamen University

\textbf{*Equal Contributions} \quad
Yuzheng Cai \quad Siqi Cai \quad Yuchen Shi \quad Zihan Xu

\setcitestyle{numbers,square}

\bibliography{citation}

\newpage
\appendix
\section{Case Study}
\label{app: case study}

On math reasoning and web searching scenarios, we provide examples illustrating how Training-Free GRPO enhances ReAct~\citep{react} with DeepSeek-V3.1-Terminus and corresponding tools.

\subsection{Experience-Guided Tool-Integrated Math Reasoning}
We consider a geometric configuration with two rectangles $ABCD$ and $EFGH$ where $D,E,C,F$ are collinear in that order, and $A,D,H,G$ are concyclic. Given $BC=16$, $AB=107$, $FG=17$, $EF=184$, the task is to determine $CE$.

\begin{figure}[!htp]
\small
\begin{tcolorbox}[colback=mylightblue, colframe=youtuBlue, enhanced jigsaw, title=\textbf{Tool-integrated Math Reasoning (without experiences)}]
\textbf{\textcolor{youtuBlue}{Question}} \\
Rectangles $ABCD$ and $EFGH$ are drawn such that $D,E,C,F$ are collinear. Also, $A,D,H,G$ all lie on a circle. If $BC=16$,$AB=107$,$FG=17$, and $EF=184$, what is the length of $CE$? \\
\tcblower
\textbf{\textcolor{youtuBlue}{Trajectory}} \\
\agentitem{I'll place point $D$ at the origin $(0,0)$ and use a horizontal line for $D,E,C,F$. For rectangle $ABCD$, set $A=(0,16)$, $B=(107,16)$, $C=(107,0)$, $D=(0,0)$.}
\toolitem{Set coordinates accordingly.}
\agentitem{Because $EFGH$ is a rectangle with $EF$ horizontal and $FG=17$, set $F=(x_f,0)$, $E=(x_f-184,0)$, and initially set $G=(x_f, \textcolor{myRed}{+17})$, $H=(x_f-184, \textcolor{myRed}{+17})$.}
\agentitem{Impose concyclicity of $A(0,16), D(0,0), G(x_f,17), H(x_f-184,17)$ using the 4-point determinant condition, derive an equation in $x_f$.}
\toolitem{Solve the resulting quadratic for $x_f$; obtain $x_f\approx184.092$ and thus $x_e=x_f-184\approx0.092$.}
\agentitem{Compute $CE=|107-x_e|\approx106.9076$.}
\agentitem{Swap point orderings ($D$–$C$–$E$–$F$, $E$–$D$–$C$–$F$) and re-derive. At one point, use an angle-sine relation and get $x_e(x_e+184)=17$ leading to $x_e=-92+\sqrt{8481}\approx0.0923$, hence $CE\approx106.9076$.}
\agentitem{Present an “exact form” $CE=199-\sqrt{8481}$ as the final answer.}
\\
\textbf{\textcolor{youtuBlue}{Outcome}} \\
Incorrect: $CE\approx106.9076$ (reported as $199-\sqrt{8481}$).
\end{tcolorbox}
\caption{Example trajectory without experience in Math Reasoning.}
\label{fig:mathcase_wo_exp}
\end{figure}

\textbf{Baseline (without experiences).}
As shown in Figure~\ref{fig:mathcase_wo_exp}, the unassisted agent initializes a coordinate system with $D=(0,0)$ and models $ABCD$ as axis-aligned. For $EFGH$, it places $E=(x_f-184,0)$, $F=(x_f,0)$ and, critically, sets $G=(x_f,\,+17)$, $H=(x_f-184,\,+17)$, i.e., with a positive vertical orientation for the short side. It then enforces the four-point concyclicity of $A(0,16)$, $D(0,0)$, $G$, $H$ via a determinant condition and solves for $x_f$, yielding $x_f\approx 184.092$ and consequently $x_e=x_f-184\approx 0.092$. From this, it reports $CE\approx 106.9076$ and an ``exact'' expression $199-\sqrt{8481}$.

This trajectory exhibits three systemic issues: (i) misinterpretation of the vertical orientation (wrong sign for the $y$-coordinates of $G,H$), (ii) inconsistent handling of the order $D$–$E$–$C$–$F$ and the lack of a unified parameterization for segment relations, and (iii) absence of systematic, comprehensive post-solution verification—i.e., no integrated check that the final coordinates simultaneously satisfy rectangle dimensions. These issues lead to an incorrect cyclic constraint (e.g., an intermediate relation of the form $x(x+184)=17$) and acceptance of a spurious solution without full geometric verification. Note that although $CE\approx 106.91$ lies within $0<CE<107$, this alone does not validate the solution; the critical failure was the lack of holistic consistency checks across all problem constraints.

\begin{figure}[!htp]
\small
\begin{tcolorbox}[colback=mylightblue, colframe=youtuBlue, enhanced jigsaw, title=\textbf{Tool-integrated Math Reasoning (with experiences)}]
\textbf{\textcolor{youtuBlue}{Question}} \\
Rectangles $ABCD$ and $EFGH$ are drawn such that $D,E,C,F$ are collinear. Also, $A,D,H,G$ all lie on a circle. If $BC=16$,$AB=107$,$FG=17$, and $EF=184$, what is the length of $CE$? \\
\\
\textbf{\textcolor{youtuBlue}{Experiences}} \\
\textnormal{[1]}  When solving geometry problems with intersections, validate solutions lie within bounded regions or segments, not on extensions, to avoid extraneous answers. \\
$\cdots$\\
\textnormal{[7]}  When solving geometry with constraints (e.g., trapezoids), validate all conditions: parallel sides via slopes, convexity via cross products, to avoid extraneous solutions.\\
$\cdots$\\
\textnormal{[29]} When constructing points on line extensions, carefully interpret directional phrases like 'beyond B'. \\
$\cdots$\\
\textnormal{[37]} In geometry problems with points on sides of a triangle and given segment lengths, first determine all three side lengths by summing the appropriate segments. For example, if points on AB, AC, BC are given with AM, MB, AN, NC, BP, PC, then AB = AM + MB, AC = AN + NC, BC = BP + PC. This ensures the triangle is consistently defined before coordinate placement or area calculations. \\
$\cdots$ \\
\tcblower
\textbf{\textcolor{youtuBlue}{Trajectory}} \\
\agentitem{\textcolor{youtuBlue}{Fix the order $D$–$E$–$C$–$F$ on a horizontal line. Set $CE=x$ with the segment constraint $0<x<107$}. (By [G29],[G1])}
\toolitem{Record relations: $DE=107-x$ and $CF=184-x$ (from [G37]).}
\toolitem{Place $D=(0,0)$, $E=(107-x,0)$, $C=(107,0)$, $F=(291-x,0)$, ensuring $EF=184$.}
\agentitem{\textcolor{youtuBlue}{Choose the vertical orientation consistently: since $A=(0,16)$ and $D=(0,0)$, set the short side $FG=17$ downward, i.e., $H=(107-x,-17)$ and $G=(291-x,-17)$}. (This corrects the sign error in the no-experience path)}
\agentitem{Impose concyclicity using the circle equation $x^2+y^2+Dx+Ey+F=0$. With $A(0,16)$ and $D(0,0)$, get $F=0$, $E=-16$.}
\toolitem{Substitute $H(107-x,-17)$ and $G(291-x,-17)$ to obtain two equations. Subtract to eliminate constants and get $D=2x-398$.}
\agentitem{Substitute back to get the quadratic $x^2-398x+30576=0.$}
\toolitem{Compute the discriminant: $398^2-4\cdot30576=36100=190^2$. Roots: $x=294$ and $x=104$.}
\agentitem{\textcolor{youtuBlue}{Apply the segment constraint $0<x<107$ to discard $x=294$ and select $x=104$} (by [G1])}
\agentitem{(\textcolor{youtuBlue}{Full verification by [G7]}) $DE=107-104=3$, $CF=184-104=80$, $EF=184$, $FG=17$. The circle $x^2+y^2-190x-16y=0$ passes through $A=(0,16),D=(0,0),H=(3,-17),G=(187,-17)$. All constraints satisfied.}
\\
\\
\textbf{\textcolor{youtuBlue}{Outcome}} \\
Correct: $CE=104$. 

\end{tcolorbox}
\caption{Example trajectory with experiences in Math Reasoning.}
\label{fig:mathcase}
\end{figure}

\textbf{Enhanced (With Experiences).}
Refer to Figure~\ref{fig:mathcase}, with a curated experience pool, the agent follows a structured pipeline:
\begin{itemize}[leftmargin=*]
    \item Directional ordering ([29]) and boundedness validation ([1]): It fixes the order $D$–$E$–$C$–$F$ on a line and sets $CE=x$ with $0<x<107$, ensuring $E$ lies on segment $DC$ and $F$ lies beyond $C$.
    \item Segment-addition parameterization ([37]): It uses $DE+EC=DC=AB=107$ and $EC+CF=EF=184$ to obtain $DE=107-x$, $CF=184-x$, and places $D=(0,0)$, $E=(107-x,0)$, $C=(107,0)$, $F=(291-x,0)$.
    \item Consistent vertical orientation and cyclic modeling: Noting $A=(0,16)$, $D=(0,0)$, it orients the short side downward ($FG=17$) so $H=(107-x,-17)$, $G=(291-x,-17)$. Using the circle equation $x^2+y^2+Dx+Ey+F=0$ with $A$ and $D$ yields $F=0$, $E=-16$. Substituting $H$ and $G$, subtracting the two equations gives $D=2x-398$; back-substitution reduces to the quadratic$x^2 - 398x + 30576 = 0$, with discriminant $398^2 - 4\cdot 30576 = 36100 = 190^2$ and roots $x=104,\,294$.
    \item Root selection and full verification ([1], [7]): Applying $0<x<107$ filters out $x=294$, selecting $x=104$. The agent then verifies all constraints: $DE=107-104=3$, $CF=184-104=80$, $EF=184$, $FG=17$, and confirms that the circle $x^2+y^2-190x-16y=0$ passes through $A=(0,16)$, $D=(0,0)$, $H=(3,-17)$, $G=(187,-17)$.
\end{itemize}

\textbf{Comparative Analysis.}
This case reveals a clear causal link between experience-guided behaviors and correctness. Experience [29] eliminates directional ambiguity and enforces the correct collinearity order, directly addressing the baseline’s misplacement of $G,H$. Experience [37] induces a clean single-variable parameterization ($DE=107-x$, $CF=184-x$), which simplifies the cyclic constraint to a solvable quadratic. Experience [1] imposes a necessary boundedness filter ($0<x<107$) to discard extraneous roots. Finally, experience [7] mandates comprehensive post-solution verification (rectangle dimensions, collinearity, concyclicity), preventing acceptance of spurious solutions.

Compared to the unassisted trajectory, the experience-informed reasoning corrects the vertical orientation, resolves ordering and parameterization inconsistencies, and installs principled validation gates. This case demonstrates the positive impact of integrating domain-specific experiences on reliability and accuracy in tool-integrated mathematical reasoning.

\subsection{Experience-Guided Web Searching}

We consider a web searching task from WebWalkerQA: quantify 2024 rewards for (i) creators in the Creator Program (weekly amount), and (ii) players in the Play-2-Airdrop during Private Beta 4 (total pool and duration).

\textbf{Baseline (without experiences).}
As summarized in Figure~\ref{fig:webscase_no_exp}, the unassisted agent issues multiple broad searches and relies heavily on result snippets and third-party summaries, delaying clicks into authoritative pages. It eventually opens the PB4 registration post to confirm “Play-2-Airdrop Campaign: 200{,}000 ILV over six months,” but continues to scan snippets for the Creator Program value without opening the relevant official post. Connection errors to one official page cause the agent to revert to broad searches rather than alternative primary-source strategies (e.g., site-specific queries or adjacent official posts). The trajectory remains incomplete: it reports the Play-2-Airdrop figure but fails to confirm the Creator Program’s “100 ILV weekly” from an official source, yielding an incorrect/incomplete answer.

\textbf{Enhanced (With Experiences).}
Refer to Figure~\ref{fig:webscase}. With a curated experience pool, the agent follows a disciplined pipeline: (1) prioritize official sources ([14], [4]) and open the PB4 overview and registration posts to extract the “200{,}000 ILV over six months” and Testnet/Mainnet allocation structure; (2) refine search terms to target formal titles ([2], [12], [1]) and open “The Illuvium Creator Program Is Getting An Upgrade!”; (3) extract the exact line “The prize pool of 100 ILV tokens awaits participating Creators each week,” and (4) synthesize both verified statements into a complete answer aligned with the question requirements ([25]). This results in the correct, fully supported output: creators earn 100 ILV weekly; players have a 200{,}000 ILV pool distributed over six months in PB4’s Play-2-Airdrop.

\begin{figure}[!htp]
\small
\begin{tcolorbox}[colback=mylightblue, colframe=youtuBlue, enhanced jigsaw, title=\textbf{Web Searching (without experiences)}]
\textbf{\textcolor{youtuBlue}{Question}} \\
In 2024, how much ILV reward can creators participating in the Creator Program and players engaging in the Play-2-Airdrop during Private Beta 4 on the Illuvium platform earn? \\
\tcblower
\textbf{\textcolor{youtuBlue}{Trajectory}} \\
\agentitem{I'll start with a broad web search for both program names and rewards in 2024.}
\toolitem{google\_search("Illuvium Creator Program Play-2-Airdrop Private Beta 4 2024 rewards ILV")}
\agentitem{From snippets, PB4 mentions a “Play-2-Airdrop campaign” and “200,000 ILV over six months.” This likely answers half the question. \textcolor{myRed}{(continue searching rather than opening)}}
\agentitem{I'll search specifically for the Creator Program rewards.}
\toolitem{google\_search("\"Illuvium Creator Program\" 2024 rewards ILV allocation")}
\agentitem{I see references (Binance Square, media blogs) to “100 ILV per week,” but can't confirm from illuvium.io quickly. I’ll keep searching.}
\agentitem{\textcolor{myRed}{Avoids clicking potential official announcements; keeps scanning search result snippets.}}
\agentitem{I'll fetch the PB4 registration page to confirm the Play-2-Airdrop details.}
\toolitem{get\_content("https://illuvium.io/news/register-now-for-private-beta-4")}
\agentitem{Extracted: “Play-2-Airdrop Campaign: 200,000 ILV tokens over six months.” That confirms the second quantity.}
\agentitem{I'll look for the Airdrop Season details page.}
\toolitem{google\_search("\"Illuvium Airdrop Season Details\" blog post 2024")}
\toolitem{get\_content("https://illuvium.io/news/illuvium-airdrop-season-details")}
\agentitem{Connection fails; \textcolor{myRed}{instead of trying alternative official sources or site-specific queries, returns to broad searches.}}
\agentitem{More searches for “Illuvium 100 ILV per week creator rewards April 2024” and “site:illuvium.io Creator Program 100 ILV” yield sparse official confirmation; I’ll defer the creator number.}
\agentitem{Fails to open likely official Creator Program pages; \textcolor{myRed}{leaves the creator number unconfirmed.}}
\agentitem{Synthesis: I will answer with what’s confirmed officially (Play-2-Airdrop 200,000 ILV over 6 months) and note the creator reward as unclear.}
\\
\\
\textbf{\textcolor{youtuBlue}{Outcome}} \\
Creators: (not clearly specified; mentions unverified “100 ILV per week”). Players (PB4 Play-2-Airdrop): 200,000 ILV over six months.
\end{tcolorbox}
\caption{Example trajectory without experiences in Web Searching.}
\label{fig:webscase_no_exp}
\end{figure}

\begin{figure}[!htp]
\small
\begin{tcolorbox}[colback=mylightblue, colframe=youtuBlue, enhanced jigsaw, title=\textbf{Web Searching (with experiences)}]
\textbf{\textcolor{youtuBlue}{Question}} \\
In 2024, how much ILV reward can creators participating in the Creator Program and players engaging in the Play-2-Airdrop during Private Beta 4 on the Illuvium platform earn? \\
\\
\textbf{\textcolor{youtuBlue}{Experiences}} \\
\textnormal{[1]} Document identification: Focus on extracting formal titles and collection names from official metadata and headers rather than inferring relationships from content descriptions. \\
$\cdots$\\
\textnormal{[2]} Systematic exploration: Progress from broad initial queries to targeted content extraction using adaptive search strategies that balance discovery with precision. \\
$\cdots$\\
\textnormal{[4]} Source prioritization: Prioritize systematic extraction from authoritative comprehensive documents over fragmented information for coherent topic coverage. \\
$\cdots$\\
\textnormal{[12]} Iterative refinement: Continuously refine search terms based on emerging patterns while periodically re-evaluating previously encountered information to avoid overlooking correct solutions and maintain essential contextual elements. \\
$\cdots$\\
\textnormal{[14]} Primary source prioritization: Prioritize locating official primary source documents over secondary coverage to ensure accuracy and capture exact language for definitive metadata and hierarchical relationships. \\
\textnormal{[25]} Partial match evaluation: Systematically evaluate partial matches from initial searches against all criteria before pursuing alternative paths. \\
$\cdots$\\
\tcblower
\textbf{\textcolor{youtuBlue}{Trajectory}} \\
\agentitem{\textcolor{youtuBlue}{Start with broad search, then pivot to official illuvium.io pages for PB4 and Creator Program.} (By [2], [14], [4])}
\toolitem{google\_search("Illuvium Creator Program rewards 2024 Private Beta 4 ILV earnings", num\_results=5)}
\agentitem{\textcolor{youtuBlue}{Identify official pages in results: “Illuvium Beta 4: All You Need to Know” and “Register Now for Private Beta 4”. Open them to confirm Play-2-Airdrop details.} (By [14], [25])}
\toolitem{get\_content("https://illuvium.io/news/illuvium-beta-4-all-you-need-to-know")}
\agentitem{Extracted from official PB4 overview: Testnet phases get 10\% of the Play-2-Airdrop pool; clarifies structure and timeline.}
\toolitem{get\_content("https://illuvium.io/news/register-now-for-private-beta-4")}
\agentitem{Extracted explicit official number: “Play-2-Airdrop Campaign: 200,000 ILV tokens” over six months. This covers the player reward for PB4 context.}
\agentitem{\textcolor{youtuBlue}{Iteratively refine search for the Creator Program exact weekly reward using formal title cues like “Creator Program … upgrade”.} (key behavior; By [12], [1])}
\toolitem{google\_search("\"Illuvium Creator Program\" rewards ILV 2024", num\_results=5)}
\agentitem{\textcolor{youtuBlue}{  Click the official post: “The Illuvium Creator Program Is Getting An Upgrade!”.} (By [14], [1], [4])}
\toolitem{get\_content("https://illuvium.io/news/illuvium-creator-program-is-getting-an-upgrade")}
\agentitem{Extracted exact statement from official source: “The prize pool of 100 ILV tokens awaits participating Creators each week.” This provides the creator reward.}
\\
\textbf{\textcolor{youtuBlue}{Outcome}} \\
Creators can earn 100 ILV weekly in the Creator Reward Program, and 200,000 ILV will be distributed over six months for players in the Play-2-Airdrop during Private Beta 4.
\end{tcolorbox}
\caption{Example trajectory with experiences in Web Searching.}
\label{fig:webscase}
\end{figure}

\textbf{Comparative Analysis.}
Experience-guided behaviors directly address baseline deficiencies: primary source prioritization ([14], [4]) removes reliance on snippets and third-party coverage; document identification ([1]) and iterative refinement ([2], [12]) ensure the agent locates and opens the exact Creator Program post; partial match evaluation ([25]) steers the agent to confirm numerical claims at their authoritative origin. In contrast, the baseline wastes context on searches without content acquisition, leaves critical values unverified, and produces an incomplete answer.

\section{Prompts}
\label{app:prompts}

In this appendix, we provide the prompts used in math reasoning tasks.
Figures~\ref{fig: math system prompt} and \ref{fig: math experience prompt} present the prompts for solving math tasks.
Figures~\ref{fig: math prompt summary} and \ref{fig: math prompt advantage} are used in group relative semantic advantage.
Figure~\ref{fig: math prompt optimization} works for optimizing the experiential knowledge $\mathcal{E}$.

\begin{figure}[!htp]
\small
\vspace{2mm}
\begin{tcolorbox}[colback=mylightblue, colframe=youtuBlue, enhanced jigsaw, title=\textbf{System Prompt}]
Solve the following problem step by step. You now have the ability to selectively write executable Python code to enhance your reasoning process, e.g., calulating numbers and verifying math computations. Never directly just printing your semantic reasoning in Python. The Python code will be executed by an external sandbox, and the output (returned as a dict with the message in the "message" field) can be returned to aid your reasoning and help you arrive at the final answer. The Python code should be complete scripts, including necessary imports. 
\vspace{2mm}

Each code snippet is wrapped with 

\textasciigrave \textasciigrave \textasciigrave python

code snippet

\textasciigrave \textasciigrave \textasciigrave.
\vspace{2mm}

The last part of your final response should be in the following format:

$<$answer$>$
$\backslash$boxed\{The final answer goes here.\}
$<$/answer$>$
\end{tcolorbox}
\vspace{-2mm}
\caption{System prompt for ReAct framework in Math Reasoning.}
\label{fig: math system prompt}
\vspace{3mm}
\end{figure}

\begin{figure}[!htp]
\small
\begin{tcolorbox}[colback=mylightblue, colframe=youtuBlue, enhanced jigsaw, title=\textbf{Prompt for Involving Experiential Knowledge $\mathcal{E}$}]
Please solve the problem:\\
\textcolor{youtuBlue}{\texttt{\{problem\}}}
\vspace{2mm}

When solving problems, you MUST first carefully read and understand the helpful instructions and experiences:\\
\textcolor{youtuBlue}{\texttt{\{experiences\}}}
\end{tcolorbox}
\vspace{-2mm}
\caption{Prompt for supplementing math problems with experiential knowledge $\mathcal{E}$.}
\label{fig: math experience prompt}
\vspace{3mm}
\end{figure}

\begin{figure}[!htp]
\small
\begin{tcolorbox}[colback=mylightblue, colframe=youtuBlue, enhanced jigsaw, title=\textbf{Prompt for Trajectory Summarization in Group Advantage Computation}]
An agent system may be provided with some experiences, and then it produces the following trajectory to solve the given problem. Please summarize the trajectory step-by-step:

1. For each step, describe what action is being taken, and which experience has been used in this step.

2. Given the grading of this rollout and the correct answer, identify and explain any steps that represent detours, errors, or backtracking, highlighting why they might have occurred and what their impact was on the trajectory's progress. 

3. Maintain all the core outcome of each step, even if it was part of a flawed process.
\vspace{2mm}

$<$trajectory$>$
\textcolor{youtuBlue}{\texttt{\{trajectory\}}}
$<$/trajectory$>$

$<$evaluation$>$
\textcolor{youtuBlue}{\texttt{\{whether the answer is correct or not\}}}
$<$/evaluation$>$

$<$groundtruth$>$
\textcolor{youtuBlue}{\texttt{\{the ground truth answer\}}}
$<$/groundtruth$>$

\vspace{2mm}
Only return the trajectory summary of each step, e.g.,

1. what happened in the first step and the core outcomes

2. what happened in the second step and the core outcomes

3. ...
\end{tcolorbox}
\vspace{-2mm}
\caption{Prompt for summarizing each trajectory in Math Reasoning.}
\label{fig: math prompt summary}
\vspace{3mm}
\end{figure}

\begin{figure}[!htp]
\small
\begin{tcolorbox}[colback=mylightblue, colframe=youtuBlue, enhanced jigsaw, title=\textbf{Prompt for Group Advantage Computation}]
An agent system is provided with a set of experiences and has tried to solve the problem multiple times with both successful and wrong solutions. Review these problem-solving attempt and extract generalizable experiences. Follow these steps:
\vspace{2mm}

1. Trajectory Analysis:

$\quad-$ For successful steps: Identify key correct decisions and insights
    
$\quad-$ For errors: Pinpoint where and why the reasoning went wrong
    
$\quad-$ Note any important patterns or strategies used/missed
    
$\quad-$ Review why some trajectories fail? Is there any existing experiences are missed, or experiences do not provide enough guidance?
\vspace{2mm}

2. Update Existing Experiences

$\quad-$ Some trajectories may be correct and others may be wrong, you should ensure there are experiences can help to run correctly
    
$\quad-$ You have three options: [modify, add, delete]
    
$\quad \quad *$ modify: You can modify current experiences to make it helpful
        
$\quad \quad *$ add: You can introduce new experiences to improve future performance

$\quad \quad *$ delete: You can delete existing experiences

$\quad-$ You can update at most \textcolor{youtuBlue}{\texttt{\{max number of operations\}}} clear, generalizable lessons for this case
    
$\quad-$ Before updating each experience, you need to:
    
$\quad \quad *$ Specify when it would be most relevant
        
$\quad \quad *$ List key problem features that make this experience applicable
        
$\quad \quad *$ Identify similar problem patterns where this advice applies
\vspace{2mm} 

3. Requirements for each experience that is modified or added.

$\quad-$ Begin with general background with several words in the experience
    
$\quad-$ Focus on strategic thinking patterns, not specific calculations
    
$\quad-$ Emphasize decision points that could apply to similar problems
\vspace{2mm}

Please provide reasoning in details under the guidance of the above 3 steps.
After the step-by-step reasoning, you will finish by returning in this JSON format as follows:

\textasciigrave \textasciigrave \textasciigrave json\\
$[$

$\quad$ \{

$\quad \quad$ "option": "modify",

$\quad \quad$ "experience": "the modified experience",

$\quad \quad$ "modified\_from": "G17" \# specify the ID of experience that is modified

$\quad$ \},
    
$\quad$ \{

$\quad \quad$ "option": "add",

$\quad \quad$ "experience": "the added experience",

$\quad$ \},
    
$\quad$ \{
    
$\quad \quad$ "option": "delete",

$\quad \quad$ "delete\_id": "the deleted experience ID",

$\quad$ \},  ...
    
$]$\\
\textasciigrave \textasciigrave \textasciigrave
\vspace{2mm}

Note that your updated experiences may not need to cover all the options. You can only use one type of updates or choose to remain all experiences unchanged.
\vspace{2mm}

$<$problem$>$
\textcolor{youtuBlue}{\texttt{\{problem\}}}
$<$/problem$>$

$<$trajectories$>$
\textcolor{youtuBlue}{\texttt{\{$G$ summarized trajectories in the same group\}}}
$<$/trajectories$>$

$<$groundtruth$>$
\textcolor{youtuBlue}{\texttt{\{answer\}}}
$<$/groundtruth$>$

$<$experience$>$
\textcolor{youtuBlue}{\texttt{\{experiences\}}}
$<$/experience$>$
\end{tcolorbox}
\caption{Prompt for group advantage computation in Math Reasoning.}
\label{fig: math prompt advantage}
\end{figure}

\begin{figure}[!htp]
\small
\begin{tcolorbox}[colback=mylightblue, colframe=youtuBlue, enhanced jigsaw, title=\textbf{Prompt for the Optimization Step}]
An agent system is provided with a set of experiences and has tried to solve the problem multiple times. From the reflections, some suggestions on the existing experiences have been posed. Your task is to collect and think for the final experience revision plan. Each final experience must satisfy the following requirements\\
1. It must be clear, generalizable lessons for this case, with no more than 32 words\\
2. Begin with general background with several words in the experience\\
3. Focus on strategic thinking patterns, not specific calculations\\
4. Emphasize decision points that could apply to similar problems\\
5. Avoid repeating saying similar experience in multiple different experiences\\
\vspace{2mm}

$<$experience$>$
\textcolor{youtuBlue}{\texttt{\{experiences\}}}
$<$/experience$>$

$<$suggested\_updates$>$
\textcolor{youtuBlue}{\texttt{\{all group advantage in current batch\}}}
$<$/suggested\_updates$>$
\vspace{2mm}

Please provide reasoning in each of the suggestions, and think for how to update existing experiences 
You have three update options: [modify, merge, delete]\\
- modify: You can modify current experiences to make it helpful\\
- merge: You can merge some similar experiences into a more general forms to reduce duplication\\
- delete: You can delete an experience
\vspace{2mm}

After generating the step-by-step reasoning, you need to give the final experience revision details by returning in this JSON format as follows:\\
\textasciigrave \textasciigrave \textasciigrave json\\
$[$

$\quad$ \{

$\quad \quad$ "option": "modify",

$\quad \quad$ "experience": "the modified experience",

$\quad \quad$ "modified\_from": "G17" \# specify the ID of experience that is modified

$\quad$ \},
    
$\quad$ \{

$\quad \quad$ "option": "merge",

$\quad \quad$ "experience": "the merged experience",

$\quad \quad$ "merged\_from": ["C1", "C3", "S4", ...] \# specify the str IDs of experiences that is merged from, at least 2 IDs are needed

$\quad$ \},
    
$\quad$ \{
    
$\quad \quad$ "option": "delete",

$\quad \quad$ "delete\_id": "the deleted experience ID",

$\quad$ \},  ...
    
$]$\\
\textasciigrave \textasciigrave \textasciigrave
\vspace{2mm}

Note that your updated experiences may not need to cover all the options. You can only use one type of updates or choose to remain all experiences unchanged.
\end{tcolorbox}
\caption{Prompt for optimizing experiential knowledge $\mathcal{E}$ based on group advantages in the same batch in Math Reasoning.}
\label{fig: math prompt optimization}
\end{figure}

\section{Examples of Learned Experiences}
\label{app: learned experiences}

\begin{figure}[!htp]
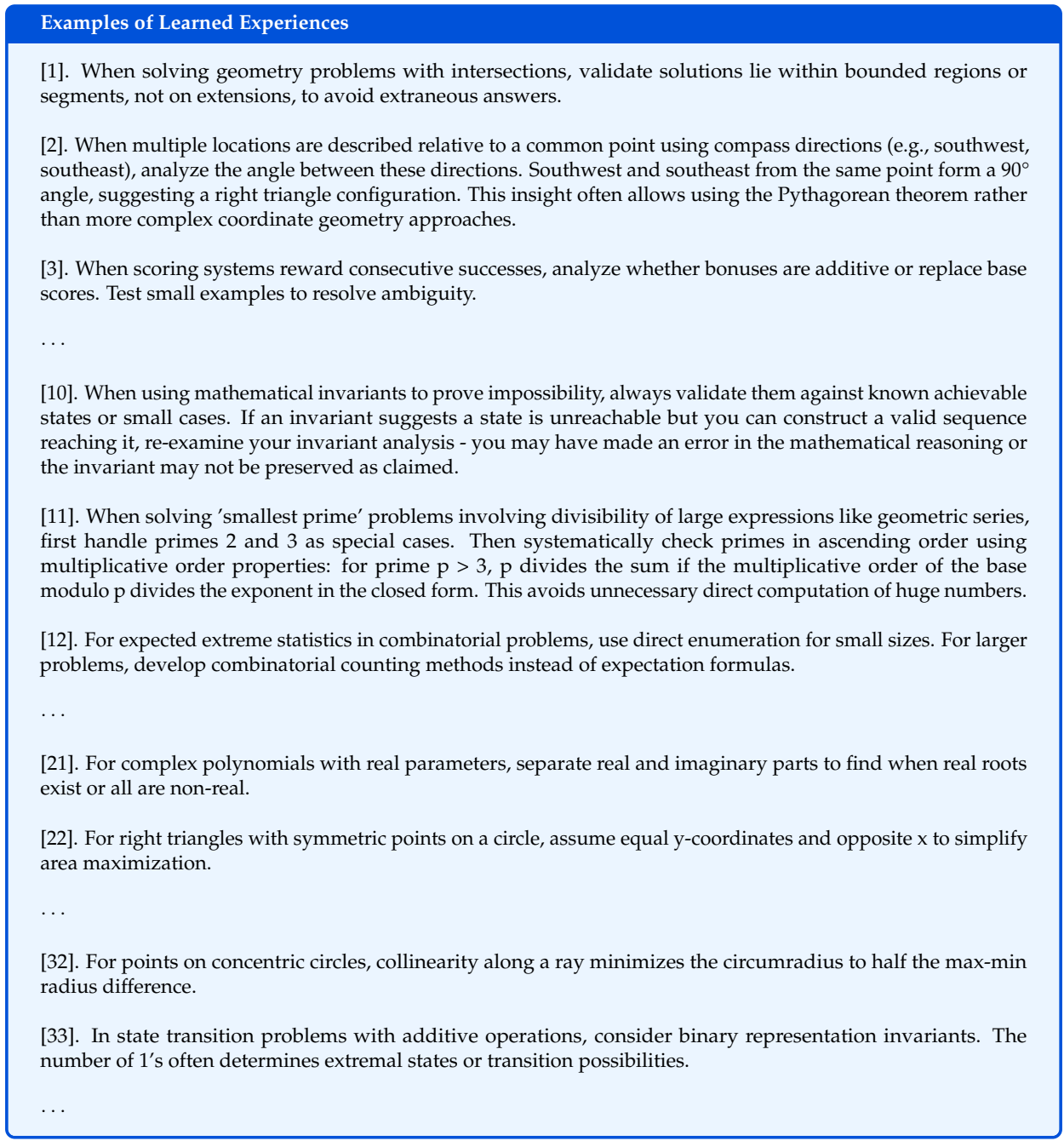

\small
\vspace{2mm}
\begin{tcolorbox}[colback=mylightblue, colframe=youtuBlue, enhanced jigsaw, title=\textbf{Examples of Learned Experiences}]
[1]. When solving geometry problems with intersections, validate solutions lie within bounded regions or segments, not on extensions, to avoid extraneous answers.\\

[2]. When multiple locations are described relative to a common point using compass directions (e.g., southwest, southeast), analyze the angle between these directions. Southwest and southeast from the same point form a 90° angle, suggesting a right triangle configuration. This insight often allows using the Pythagorean theorem rather than more complex coordinate geometry approaches.\\

[3]. When scoring systems reward consecutive successes, analyze whether bonuses are additive or replace base scores. Test small examples to resolve ambiguity.\\

$\cdots$\\

[10]. When using mathematical invariants to prove impossibility, always validate them against known achievable states or small cases. If an invariant suggests a state is unreachable but you can construct a valid sequence reaching it, re-examine your invariant analysis - you may have made an error in the mathematical reasoning or the invariant may not be preserved as claimed.\\

[11]. When solving 'smallest prime' problems involving divisibility of large expressions like geometric series, first handle primes 2 and 3 as special cases. Then systematically check primes in ascending order using multiplicative order properties: for prime p > 3, p divides the sum if the multiplicative order of the base modulo p divides the exponent in the closed form. This avoids unnecessary direct computation of huge numbers.\\

[12]. For expected extreme statistics in combinatorial problems, use direct enumeration for small sizes. For larger problems, develop combinatorial counting methods instead of expectation formulas.\\

$\cdots$\\

[21]. For complex polynomials with real parameters, separate real and imaginary parts to find when real roots exist or all are non-real.\\

[22]. For right triangles with symmetric points on a circle, assume equal y-coordinates and opposite x to simplify area maximization.\\

$\cdots$\\

[32]. For points on concentric circles, collinearity along a ray minimizes the circumradius to half the max-min radius difference.\\

[33]. In state transition problems with additive operations, consider binary representation invariants. The number of 1's often determines extremal states or transition possibilities.\\

$\cdots$
\end{tcolorbox}
\caption{Example of learned experiences from Training-Free GRPO with tool use in Math Reasoning.}
\label{fig:learned experiences}
\end{figure}

In this appendix, we provide some examples in Figure~\ref{fig:learned experiences}, which are extracted from 48 learned experiences by Training-Free GRPO with tool use in Math Reasoning.

\end{document}